\def\arxiv{true}
\pdfoutput=1

\documentclass[11pt]{article}

\ifx\arxiv\undefined
    \usepackage[review]{acl}
\else
    \usepackage[final]{acl}
\fi

\usepackage{times}
\usepackage{latexsym}

\usepackage[T1]{fontenc}

\usepackage[utf8]{inputenc}

\usepackage{microtype}

\usepackage{inconsolata}

\usepackage{graphicx}
\usepackage{tcolorbox}
\usepackage{xcolor}
\tcbset{colframe=blue!75!black, colback=blue!10!white, boxrule=0.5mm, arc=3mm}
\usepackage{hyperref}

\usepackage{amsmath}
\usepackage{amssymb}
\usepackage{mathtools}
\usepackage{amsthm}

\usepackage{booktabs, multirow, makecell} 
\usepackage{soul}

\input{additional_imports}

\usepackage[utf8]{inputenc}
\usepackage{hyphenat}
\usepackage{xspace}
\usepackage{amsmath}
\usepackage{amsfonts}
\usepackage{hyperref}
\usepackage{url}
\usepackage{booktabs}
\usepackage{multirow}
\usepackage{makecell}
\usepackage{minibox}
\usepackage{bbm}
\usepackage{graphicx}
\usepackage{balance}
\usepackage{mathtools}
\usepackage{color}
\usepackage{marvosym}
\usepackage{ifthen}
\usepackage{textcomp}
\usepackage{enumitem}
\usepackage{verbatim}
\usepackage{algorithm}
\usepackage{algorithmic}
\usepackage{numprint}
\usepackage{balance}

\newcommand{\chatoDisplayMode}[1]{#1}

\definecolor{MyRed}{rgb}{0.6,0.0,0.0} 
\definecolor{MyBlack}{rgb}{0.1,0.1,0.1} 
\newcommand{\inred}[1]{{\color{MyRed}\sf\textbf{\textsc{#1}}}}
\newcommand{\frameit}[2]{
  \begin{center}
  {\color{MyRed}
  \framebox[.9\columnwidth][l]{
    \begin{minipage}{.85\columnwidth}
    \inred{#1}: {\sf\color{MyBlack}#2}
    \end{minipage}
  }\\
  }
  \end{center}
}

\newcommand{\note}[2][]{\chatoDisplayMode{\def\@tmpsig{#1}\frameit{{\Pointinghand} Note}{#2\ifx \@tmpsig \@empty \else \mbox{ --\em #1}\fi}}}
\newcommand{\todo}[2][]{\chatoDisplayMode{\def\@tmpsig{#1}\frameit{{\Writinghand} To-do}{#2\ifx \@tmpsig \@empty \else \mbox{ --\em #1}\fi}}}

\newcommand{\abbrevStyle}[1]{#1}

\newcommand{\eg}{\abbrevStyle{e.g.}\xspace}

\newcommand{\xhdr}[1]{\vspace{1.7mm}\noindent{{\bf #1.}}}

\newcommand{\textcite}[1]{\citeauthor{#1} \shortcite{#1}}

\newcommand{\mycaption}[2]{\caption[#1]{\textbf{#1.} #2}}

\usepackage{etoolbox}

\definecolor{mycolor_blue}{HTML}{E7EFFA}
\definecolor{mycolor_green}{HTML}{E6F8E0}
\definecolor{mycolor_gray}{HTML}{ECECEC}
\definecolor{pearDark}{HTML}{2980B9}

\newcommand{\figcaption}[1]{\vspace*{-3mm}\caption{#1}\vspace*{-5mm}}
\newcommand{\moveup}{\vspace*{-2mm}}
\newcommand{\moveups}{\vspace*{-1mm}}
\newcommand{\ignore}[1]{}

\usepackage[capitalize,noabbrev]{cleveref}

\theoremstyle{plain}

\theoremstyle{definition}

\theoremstyle{remark}

\title{Are Retrials All You Need? Enhancing Large Language Model Reasoning Without Verbalized Feedback}

\author{Nearchos Potamitis \\
  Aarhus University\\
  \texttt{nearchos.potamitis@cs.au.dk} \And
  Akhil Arora \\
  Aarhus University \\
  \texttt{akhil.arora@cs.au.dk} }

\begin{document}
\maketitle

\begin{abstract}
Recent advancements in large language models (LLMs) have catalyzed the development of general-purpose autonomous agents, demonstrating remarkable performance in complex reasoning tasks across various domains. This surge has spurred the evolution of a plethora of prompt-based reasoning frameworks. A recent focus has been on iterative reasoning strategies that refine outputs through self-evaluation and verbalized feedback. However, these strategies require additional computational complexity to enable models to recognize and correct their mistakes, leading to a significant increase in their cost. In this work, we introduce the concept of \emph{``retrials without feedback''}, an embarrassingly simple yet powerful mechanism for enhancing reasoning frameworks by allowing LLMs to retry problem-solving attempts upon identifying incorrect answers. Unlike conventional iterative refinement methods, our method does not require explicit self-reflection or verbalized feedback, simplifying the refinement process. Our findings indicate that simpler retrial-based approaches often outperform more sophisticated reasoning frameworks, suggesting that the benefits of complex methods may not always justify their computational costs. By challenging the prevailing assumption that more intricate reasoning strategies inherently lead to better performance, our work offers new insights into how simpler, more efficient approaches can achieve optimal results. So, are retrials all you need?
\end{abstract}

\section{Introduction}

With strong reasoning and problem-solving abilities, large language models (LLMs)~\cite{brown2020fewshot} such as GPT-4~\cite{gpt4,gpt-4o-mini}, LLaMA~\cite{llama, llama2, llama3}, and PaLM~\cite{palm2}, have sparked a new-found interest in building general-purpose autonomous agents. LLM-based agents have portrayed excellent performance on reasoning~\cite{gsm8k} and knowledge-intensive tasks~\cite{Wikispeedia}, often requiring interactions with complex environments, such as playing complex video games~\cite{minedojo}, performing web navigation~\cite{webshop}, or enabling tool-use~\cite{toolformer}.

Naturally, the rise of LLM-based agents has contributed to the prosperity of prompt-based reasoning frameworks~\cite{cot, cot_sc, got, aot, bot, tot, lats, reflexion, react, foa} that further enhance the problem-solving and reasoning abilities of LLMs. 

\xhdr{Iterative refinement and its challenges} A burgeoning area of research in this field involves the exploration of iterative reasoning strategies that refine and improve responses through self-evaluation. Methods such as Self-Refine \cite{madaan2023selfrefine} and Reflexion \cite{reflexion} employ explicit verbalized feedback to guide an LLM in correcting its mistakes and refining its outputs. While effective, these approaches add an additional layer of complexity, requiring the model to both recognize its own errors and articulate useful self-correction strategies. Moreover, owing to an incremental increase in the context window size, verbalized feedback monotonically increases the cost for every subsequent iteration of the overall reasoning pipeline, thereby rendering iterative refinement-based methods exorbitantly expensive. For instance, with GPT-4 as the base model, RAFA~\cite{rafa}, the state-of-the-art refinement strategy, requires a \textbf{whopping $\simeq$ 600\$} on the Game of 24 benchmark task~\cite{tot} comprising just 100 examples in the test set.

\xhdr{Present work}
In this paper, we introduce the concept of \textbf{``retrials without feedback''}, an embarrassingly simple yet effective mechanism that enhances reasoning frameworks by allowing them to retry a problem-solving attempt whenever an incorrect answer is identified. Unlike Reflexion~\cite{reflexion}, which relies on explicit self-generated feedback, retrials operate without requiring verbalized introspection: an incorrect answer simply triggers another attempt until the correct solution is found or a predefined computational budget is exhausted.

Our results show that under the retrial mechanism, simpler methods such as chain-of-thoughts~\cite{cot} often outperform more sophisticated reasoning frameworks such as tree-of-thoughts~\cite{tot} or Reflexion~\cite{reflexion}. This suggests that, when given the opportunity to retry, the added complexity of sophisticated reasoning frameworks and self-reflective approaches may not always justify their computational cost. This raises a fundamental question: Are retrials all you need? By reframing the evaluation of reasoning strategies through the lens of cost efficiency, our work challenges the assumption that more intricate frameworks necessarily lead to better performance and highlights the importance of reconsidering optimization strategies for reasoning with LLMs.

\section{Related Work}

\begin{figure*}[t]
\centering
\moveup
\includegraphics[width=0.95\linewidth]{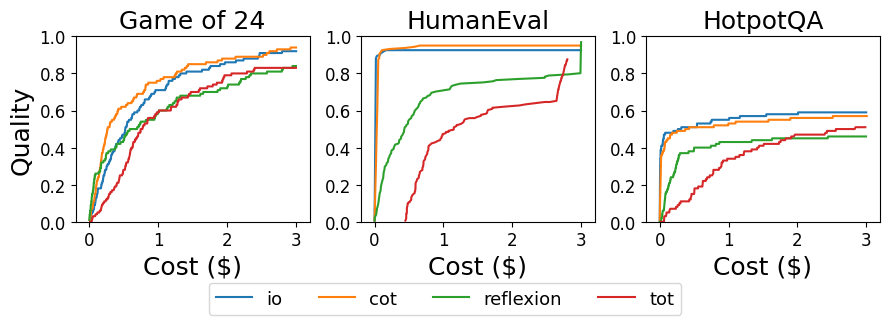}
\figcaption{Comparing the cost-quality trade-off of IO, CoT, ToT, and Reflexion using \textbf{GPT-4o-mini} as the base model. Within the indicated budget, simpler methods outperform more complex ones while remaining cost-efficient.}
\label{fig:cost_gpt}
\moveups
\end{figure*}
\label{sec:cost_gpt}

In this section, we review works that overlap closely with our study

\xhdr{Prompt-based reasoning}
Recent research focuses on developing 
strategies to enhance the reasoning capabilities of LLMs. 
Few-shot prompting employs demonstrations of high-quality input/output samples to exploit the eagerness of the LLMs to imitate patterns seen in their context window \cite{brown2020fewshot}.
Algorithm of thoughts (AoT)~\cite{aot}, goes a step further by including algorithmic examples within the prompt to propel the LLM through algorithmic reasoning pathways.
Chain-of-Thought (CoT) prompting~\cite{maxwell2021scratchpad, cot, kojima2022stepbystep} as well as other variants such as Decomposed Prompting~\cite{decomposed_prompting} and Least-to-Most \cite{least_to_most} guide LLMs to decompose a complex question into a sequence of thoughts and then synthesize an answer by resolving them methodically. 
It has been shown that Self-Consistency (CoT-SC) \cite{self_consistency} can be used to augment such methods by generating multiple thought sequences and then selecting the most accurate answer through majority voting.
Recent meta-prompting techniques \cite{meta_prompting} employ a uniform, task-independent prompting framework across multiple tasks, enabling a single LLM to iteratively refine its responses and dynamically adapt to diverse input queries.
The Buffer of Thoughts (BoT) \cite{bot} framework extracts task-specific information, uses it to retrieve relevant thought templates from its meta-buffer, and then instantiates them with more task-specific reasoning structures before continuing with the reasoning process.

\xhdr{Refinement}
Closed-loop approaches that allow an LLM to interact with an external environment can help in choosing and potentially revising an action. 
Notable examples are ReAct \cite{react}, REFINER \cite{paul2023refiner} and Self-Refine \cite{madaan2023selfrefine}. 
Reflexion \cite{reflexion} provides further linguistic feedback based on previous attempts while AdaPlanner \cite{adaplanner} also incorporates positive and negative feedback of an individual trajectory. 
Reason for future, act for now (RAFA) \cite{rafa} develops further by planning a trajectory, gathering feedback for the potential planned actions, and then revising the trajectory based on the feedback.

\xhdr{Tree search}
Thoughts are individual ideas or steps in reasoning, and when connected together, they can be modeled as a tree data structure. 
Tree search algorithms can then be used to explore a tree of thoughts and optimize the search for a final answer.
In ``Tree of Thoughts'' (ToT), the authors utilize a value function that compares different branches to describe both DFS and BFS flavors of a guided tree-search \cite{tot}.
The closely related ``Graph of Thoughts'' (GoT) approach relaxes the assumption of a strict tree structure \cite{got}.
Reasoning via Planning (RAP) \cite{rap_reasoner} augments LLMs with a world model and employs Monte Carlo Tree Search (MCTS)-based planning to reduce the search complexity. Language Agent Tree Search (LATS) \cite{lats} extends this concept by leveraging environment interactions, thereby eliminating the need for a world model.

\section{Experiments}
In this section, we provide details on the benchmarks and the different types of analyses that we performed for our experiments. Please refer to Appendix \ref{sec:appendix-methods} for more information regarding the reasoning strategies used in this study. For additional results, please see Appendix~\ref{app:additional_results}.

\subsection{Benchmark tasks}

\xhdr{Game of 24}
Game of 24 is a mathematical puzzle, where four numbers are given, and the objective is to form an arithmetic expression that equals 24 using each number exactly once. The benchmark data consists of 1362 puzzles. Following ToT~\cite{tot}, we use the puzzles indexed 901-1000 as the test set.
To evaluate the quality of the methods we use \emph{success rate}, that is, the percentage of solved puzzles. For efficiency, we use cost (in US\$).

\xhdr{HumanEval}
HumanEval is a programming puzzle that measures functional correctness for synthesizing programs from natural language docstrings. Following Reflexion \cite{reflexion}, we evaluate the methods on 161 Python programs.
We use the accuracy evaluation \emph{pass@1} to measure the quality of the benchmark. For efficiency, we use cost (in US\$).

\xhdr{HotpotQA}
HotpotQA \cite{hotpotqa} is a large-scale question-answering dataset designed to evaluate multi-hop reasoning. Multi-step methods such as ToT, are allowed to use an interactive API environment which allows the agent to search for documents and look up specific information within them. Following previous methods \cite{lats, reflexion} we evaluate on 100 randomly selected samples of their choice.
We measure the quality of the answer based on whether there is an \emph{exact match} (EM), given an oracle answer. For efficiency, we use cost (in US\$),

\subsection{Experiment setup}

\xhdr{Baselines} We analyze the impact of retrials on four prompting strategies: (1) Standard IO prompting, (2) CoT~\cite{cot}, (3) ToT~\cite{tot}, and (4) Reflexion~\cite{reflexion}. 

\xhdr{Base model} We use GPT-4o-mini~\cite{gpt-4o-mini} and LLaMA-3.3-70B~\cite{llama3} as the base models.

\subsection{Analysis}

\begin{figure*}[t]
\centering
\moveup
\includegraphics[width=0.9\linewidth]{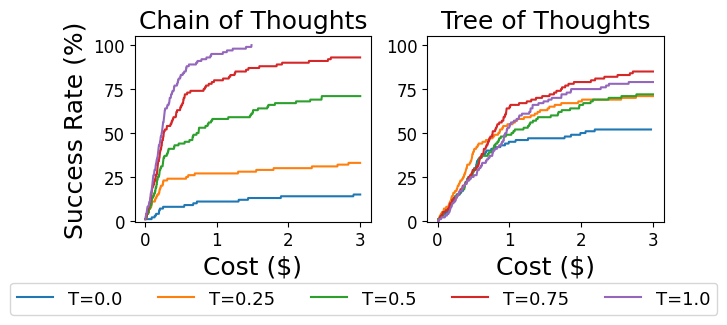}
\figcaption{Comparing the cost-quality trade-off of CoT and ToT across different temperature levels using \textbf{GPT-4o-mini} as the base model. For CoT success rate is strictly increasing as temperature increases and so does for ToT but not strictly.}
\label{fig:temp_gpt}
\moveups
\end{figure*}
\label{sec:temp_gpt}

\xhdr{Cost and Number of Retrials Analysis} This experiment aims to evaluate the effectiveness of different methods under a constrained budget. Each method follows an iterative approach, attempting to solve all of the samples in the first trial. Unsolved samples are re-tried in subsequent trials until either the budget is exhausted or of them all are solved. If the budget runs out during an iteration, the method halts immediately. To investigate cost-effectiveness we present the quality and cost for GPT-4o-mini in Fig.~\ref{fig:cost_gpt}. The corresponding plot for LLama 3.3 70B can be found in the appendix (Fig.~\ref{fig:cost_llama}). To also display sample efficiency, we plot quality and number of trials.

\xhdr{Temperature Analysis} To better understand the cost-effectiveness of each method, we repeated the previous experiments using different temperature values. Due to budget and time restrictions, we only conduct this analysis for Game of 24 across the CoT and ToT methods. The results can be found in Fig.~\ref{fig:temp_gpt} for GPT-4o-mini and Fig.~\ref{fig:temp_llama} for Llama-3.3-70B.

\section{Main results}

\xhdr{Cost efficiency}
Our results demonstrate that simple methods such as IO and CoT prompting are significantly more efficient than complex reasoning approaches such as ToT~\cite{tot} or Reflexion~\cite{reflexion}. Specifically, across all benchmarks and models, CoT prompting consistently outperforms alternative methods, often by a considerable margin. In fact, for Game of 24, CoT achieved a 94\% success rate: a result that methods such as \cite{rafa, tot} would need multiple hundreds of dollars ($\simeq$ 600\$) to achieve on GPT-4.

Cost efficiency appears to be influenced not only by the method employed but also by the specific task. For instance, in the HumanEval and HotpotQA tasks, both CoT and IO methods reach a performance plateau earlier, whereas, in the Game of 24 task, the efficiency peak is more gradual. In addition, the base model plays a crucial role in performance disparities. This is particularly evident in the quality gap between GPT-4o-mini and Llama-3.3-70B in the Game of 24 and HotpotQA benchmarks. While the two methods exhibit comparable performance with the former model, CoT substantially outperforms IO in the latter. These findings suggest that base models with stronger inherent reasoning capabilities (\eg GPT-4o-mini) can simultaneously enhance the cost-quality trade-off of simple methods such as IO prompting.

\xhdr{Temperature analysis}
For GPT-4o-mini (Fig.~\ref{fig:temp_gpt}), we can clearly see that a higher temperature value results in a better success rate. This is clearly visible for CoT as the gap between each temperature result is significantly wide. Furthermore, we can even see that for a temperature value of $1.0$ the experiment already achieves $100\%$ accuracy at half of the allocated budget. For ToT, this is also predominantly true. A possible reason why this is not as evident in the latter case is that multi-query methods such as ToT, often introduce complex prompting schemes that are affected in arbitrary ways by an increase in the temperature values. 

\section{Discussion and Concluding Insights}
\xhdr{Summary of Findings}
Our results in Figs. \ref{fig:cost_gpt} and \ref{fig:cost_llama} have shown that methods such as Chain of Thought are more cost-efficient than complex reasoning strategies on a variety of models and tasks. In fact, for some tasks, we have achieved state-of-the-art quality performance with minimal resources in terms of cost and model capabilities. Based on the results we have also shown that the cost-efficiency of a method is affected not only by the task but by the model as well. Finally, we have shown that the performance  of the retrial concept can be improved even further by properly tuning the model's temperature

\xhdr{Future work}
In the future, we would like to extend our results by showcasing the behavior of the methods when allowing an even bigger budget. For the case of HumanEval, it is already apparent that Reflexion and Tree of Thoughts will surpass the other methods if allowed for more budget. Even though this does not go against our claims regarding cost-efficiency, we would like to investigate further and showcase the full picture.
Additionally, we aim to explore methods that leverage the occurrence of retrials to further optimize reasoning processes. Specifically, we seek to develop techniques that can exploit the iterative nature of retrials to improve efficiency, potentially reducing the number of attempts needed to reach a correct solution.
Finally, our method uses trivial deterministic verifiers to decide whether an answer has been found and in this case, no retry is needed. We would aim to extend our method to tasks where there's not a trivial deterministic verifier to validate the answer.
Overall, we hope that our work will inspire further research into the role of retrials in cost-efficient reasoning and the broader optimization of reasoning frameworks for LLM-based problem-solving

\section*{Limitations}
The main limitation of our work is that it cannot be applied to tasks where the answer cannot be directly verified. For example, in the game of 24, the goal is to find a mathematical formula that evaluates to 24 given some input numbers. Once a formula is found, checking whether it equals 24 is straightforward. Similarly, in HumanEval, an answer is verified based on whether the generated function passes specific tests. This allows us to stop retrying a puzzle once a correct answer has been found. However, in cases similar to HotpotQA, the correct answer is hidden and only used for evaluation, making direct verification impossible during the solving process. As discussed in our future work, we aim to extend our method to tasks where answers cannot be deterministically verified in a trivial manner.

\section*{Acknowledgements}
We thank Chris Schwiegelshohn and Niket Tandon for insightful discussions. Arora's lab is partly supported by grants from the Novo Nordisk Foundation (NNF24OC0099109), the Pioneer Centre for AI, and EU Horizon 2020 (101168951).

\bibliography{retrial}

\begin{thebibliography}{34}
\providecommand{\natexlab}[1]{#1}

\bibitem[{Achiam et~al.(2024)Achiam, Adler, Agarwal, Ahmad, Akkaya, Aleman,
  Almeida, Altenschmidt, Altman, Anadkat et~al.}]{gpt4}
Josh Achiam, Steven Adler, Sandhini Agarwal, Lama Ahmad, Ilge Akkaya,
  Florencia~Leoni Aleman, Diogo Almeida, Janko Altenschmidt, Sam Altman,
  Shyamal Anadkat, et~al. 2024.
\newblock \href {https://arxiv.org/abs/2303.08774} {Gpt-4 technical report}.
\newblock \emph{arXiv preprint arXiv:2303.08774}.

\bibitem[{Anil et~al.(2023)Anil, Dai, Firat, Johnson, Lepikhin, Passos,
  Shakeri, Taropa, Bailey, Chen et~al.}]{palm2}
Rohan Anil, Andrew~M. Dai, Orhan Firat, Melvin Johnson, Dmitry Lepikhin,
  Alexandre Passos, Siamak Shakeri, Emanuel Taropa, Paige Bailey, Zhifeng Chen,
  et~al. 2023.
\newblock \href {https://arxiv.org/abs/2305.10403} {Palm 2 technical report}.
\newblock \emph{arXiv preprint arXiv:2305.10403}.

\bibitem[{Besta et~al.(2024)Besta, Blach, Kubicek, Gerstenberger, Podstawski,
  Gianinazzi, Gajda, Lehmann, Niewiadomski, Nyczyk et~al.}]{got}
Maciej Besta, Nils Blach, Ales Kubicek, Robert Gerstenberger, Michal
  Podstawski, Lukas Gianinazzi, Joanna Gajda, Tomasz Lehmann, Hubert
  Niewiadomski, Piotr Nyczyk, et~al. 2024.
\newblock Graph of thoughts: Solving elaborate problems with large language
  models.
\newblock In \emph{AAAI}, pages 17682--17690.

\bibitem[{Brown et~al.(2020)Brown, Mann, Ryder, Subbiah, Kaplan, Dhariwal,
  Neelakantan, Shyam, Sastry, Askell et~al.}]{brown2020fewshot}
Tom Brown, Benjamin Mann, Nick Ryder, Melanie Subbiah, Jared~D Kaplan, Prafulla
  Dhariwal, Arvind Neelakantan, Pranav Shyam, Girish Sastry, Amanda Askell,
  et~al. 2020.
\newblock Language models are few-shot learners.
\newblock In \emph{NeurIPS}, volume~33, pages 1877--1901.

\bibitem[{Cobbe et~al.(2021)Cobbe, Kosaraju, Bavarian, Chen, Jun, Kaiser,
  Plappert, Tworek, Hilton, Nakano, Hesse, and Schulman}]{gsm8k}
Karl Cobbe, Vineet Kosaraju, Mohammad Bavarian, Mark Chen, Heewoo Jun, Lukasz
  Kaiser, Matthias Plappert, Jerry Tworek, Jacob Hilton, Reiichiro Nakano,
  Christopher Hesse, and John Schulman. 2021.
\newblock \href {https://arxiv.org/abs/2110.14168} {Training verifiers to solve
  math word problems}.
\newblock \emph{arXiv preprint arXiv:2110.14168}.

\bibitem[{Fan et~al.(2022)Fan, Wang, Jiang, Mandlekar, Yang, Zhu, Tang, Huang,
  Zhu, and Anandkumar}]{minedojo}
Linxi Fan, Guanzhi Wang, Yunfan Jiang, Ajay Mandlekar, Yuncong Yang, Haoyi Zhu,
  Andrew Tang, De-An Huang, Yuke Zhu, and Anima Anandkumar. 2022.
\newblock Minedojo: Building open-ended embodied agents with internet-scale
  knowledge.
\newblock In \emph{NeurIPS: Datasets and Benchmarks Track}.

\bibitem[{Grattafiori et~al.(2024)}]{llama3}
Aaron Grattafiori et~al. 2024.
\newblock \href {https://arxiv.org/abs/2407.21783} {The llama 3 herd of
  models}.
\newblock \emph{arXiv preprint arXiv:2407.21783}.

\bibitem[{Hao et~al.(2023)Hao, Gu, Ma, Hong, Wang, Wang, and Hu}]{rap_reasoner}
Shibo Hao, Yi~Gu, Haodi Ma, Joshua~Jiahua Hong, Zhen Wang, Daisy~Zhe Wang, and
  Zhiting Hu. 2023.
\newblock Reasoning with language model is planning with world model.
\newblock In \emph{EMNLP}.

\bibitem[{Khot et~al.(2023)Khot, Trivedi, Finlayson, Fu, Richardson, Clark, and
  Sabharwal}]{decomposed_prompting}
Tushar Khot, Harsh Trivedi, Matthew Finlayson, Yao Fu, Kyle Richardson, Peter
  Clark, and Ashish Sabharwal. 2023.
\newblock \href {https://openreview.net/forum?id=_nGgzQjzaRy} {Decomposed
  prompting: A modular approach for solving complex tasks}.
\newblock In \emph{ICLR}.

\bibitem[{Kojima et~al.(2022)Kojima, Gu, Reid, Matsuo, and
  Iwasawa}]{kojima2022stepbystep}
Takeshi Kojima, Shixiang~(Shane) Gu, Machel Reid, Yutaka Matsuo, and Yusuke
  Iwasawa. 2022.
\newblock Large language models are zero-shot reasoners.
\newblock In \emph{NeurIPS}, pages 22199--22213.

\bibitem[{Liu et~al.(2024)Liu, Hu, Zhang, Guo, Ke, Liu, and Wang}]{rafa}
Zhihan Liu, Hao Hu, Shenao Zhang, Hongyi Guo, Shuqi Ke, Boyi Liu, and Zhaoran
  Wang. 2024.
\newblock Reason for future, act for now: A principled architecture for
  autonomous {LLM} agents.
\newblock In \emph{ICML}.

\bibitem[{Madaan et~al.(2023)Madaan, Tandon, Gupta, Hallinan, Gao, Wiegreffe,
  Alon, Dziri, Prabhumoye, Yang et~al.}]{madaan2023selfrefine}
Aman Madaan, Niket Tandon, Prakhar Gupta, Skyler Hallinan, Luyu Gao, Sarah
  Wiegreffe, Uri Alon, Nouha Dziri, Shrimai Prabhumoye, Yiming Yang, et~al.
  2023.
\newblock Self-refine: Iterative refinement with self-feedback.
\newblock \emph{arXiv preprint arXiv:2303.17651}.

\bibitem[{Nye et~al.(2021)Nye, Andreassen, Gur{-}Ari, Michalewski, Austin,
  Bieber, Dohan, Lewkowycz, Bosma, Luan, Sutton, and
  Odena}]{maxwell2021scratchpad}
Maxwell~I. Nye, Anders~Johan Andreassen, Guy Gur{-}Ari, Henryk Michalewski,
  Jacob Austin, David Bieber, David Dohan, Aitor Lewkowycz, Maarten Bosma,
  David Luan, Charles Sutton, and Augustus Odena. 2021.
\newblock \href {https://arxiv.org/abs/2112.00114} {Show your work: Scratchpads
  for intermediate computation with language models}.
\newblock \emph{CoRR}, abs/2112.00114.

\bibitem[{OpenAI and et~al.(2024)}]{gpt-4o-mini}
OpenAI and Hurst et~al. 2024.
\newblock \href {https://arxiv.org/abs/2410.21276} {Gpt-4o system card}.
\newblock \emph{Preprint}, arXiv:2410.21276.

\bibitem[{Paul et~al.(2023)Paul, Ismayilzada, Peyrard, Borges, Bosselut, West,
  and Faltings}]{paul2023refiner}
Debjit Paul, Mete Ismayilzada, Maxime Peyrard, Beatriz Borges, Antoine
  Bosselut, Robert West, and Boi Faltings. 2023.
\newblock \href {https://arxiv.org/abs/2304.01904} {Refiner: Reasoning feedback
  on intermediate representations}.
\newblock \emph{arXiv preprint arXiv:2304.01904}.

\bibitem[{Potamitis et~al.(2024)Potamitis, Klein, Aydin, Gulcehre, West, and
  Arora}]{foa}
Nearchos Potamitis, Lars Klein, Roland Aydin, Caglar Gulcehre, Robert West, and
  Akhil Arora. 2024.
\newblock \href {https://arxiv.org/abs/2405.06691} {Fleet of agents:
  Coordinated problem solving with large language models}.
\newblock \emph{Preprint}, arXiv:2405.06691.

\bibitem[{Schick et~al.(2023)Schick, Dwivedi-Yu, Dessi, Raileanu, Lomeli,
  Hambro, Zettlemoyer, Cancedda, and Scialom}]{toolformer}
Timo Schick, Jane Dwivedi-Yu, Roberto Dessi, Roberta Raileanu, Maria Lomeli,
  Eric Hambro, Luke Zettlemoyer, Nicola Cancedda, and Thomas Scialom. 2023.
\newblock Toolformer: Language models can teach themselves to use tools.
\newblock In \emph{NeurIPS}.

\bibitem[{Sel et~al.(2024)Sel, Tawaha, Khattar, Jia, and Jin}]{aot}
Bilgehan Sel, Ahmad Tawaha, Vanshaj Khattar, Ruoxi Jia, and Ming Jin. 2024.
\newblock Algorithm of thoughts: Enhancing exploration of ideas in large
  language models.
\newblock In \emph{ICML}.

\bibitem[{Shinn et~al.(2023)Shinn, Cassano, Gopinath, Narasimhan, and
  Yao}]{reflexion}
Noah Shinn, Federico Cassano, Ashwin Gopinath, Karthik Narasimhan, and Shunyu
  Yao. 2023.
\newblock Reflexion: language agents with verbal reinforcement learning.
\newblock In \emph{NeurIPS}, pages 8634--8652.

\bibitem[{Sun et~al.(2023)Sun, Zhuang, Kong, Dai, and Zhang}]{adaplanner}
Haotian Sun, Yuchen Zhuang, Lingkai Kong, Bo~Dai, and Chao Zhang. 2023.
\newblock Adaplanner: Adaptive planning from feedback with language models.
\newblock In \emph{NeurIPS}, volume~36, pages 58202--58245.

\bibitem[{Suzgun and Kalai(2024)}]{meta_prompting}
Mirac Suzgun and Adam~Tauman Kalai. 2024.
\newblock \href {https://arxiv.org/abs/2401.12954} {Meta-prompting: Enhancing
  language models with task-agnostic scaffolding}.
\newblock \emph{arXiv preprint arXiv:2401.12954}.

\bibitem[{Touvron et~al.(2023{\natexlab{a}})Touvron, Lavril, Izacard, Martinet,
  Lachaux, Lacroix, Rozi{\`e}re, Goyal, Hambro, Azhar et~al.}]{llama}
Hugo Touvron, Thibaut Lavril, Gautier Izacard, Xavier Martinet, Marie-Anne
  Lachaux, Timoth{\'e}e Lacroix, Baptiste Rozi{\`e}re, Naman Goyal, Eric
  Hambro, Faisal Azhar, et~al. 2023{\natexlab{a}}.
\newblock Llama: Open and efficient foundation language models.
\newblock \emph{ArXiv}, abs/2302.13971.

\bibitem[{Touvron et~al.(2023{\natexlab{b}})Touvron, Martin, Stone, Albert,
  Almahairi, Babaei, Bashlykov, Batra, Bhargava, Bhosale et~al.}]{llama2}
Hugo Touvron, Louis Martin, Kevin~R. Stone, Peter Albert, Amjad Almahairi,
  Yasmine Babaei, Nikolay Bashlykov, Soumya Batra, Prajjwal Bhargava, Shruti
  Bhosale, et~al. 2023{\natexlab{b}}.
\newblock Llama 2: Open foundation and fine-tuned chat models.
\newblock \emph{arXiv preprint arXiv:2307.09288}.

\bibitem[{Wang et~al.(2022)Wang, Wei, Schuurmans, Le, Chi, Narang, Chowdhery,
  and Zhou}]{self_consistency}
Xuezhi Wang, Jason Wei, Dale Schuurmans, Quoc Le, Ed~Chi, Sharan Narang,
  Aakanksha Chowdhery, and Denny Zhou. 2022.
\newblock Self-consistency improves chain of thought reasoning in language
  models.
\newblock \emph{arXiv preprint arXiv:2203.11171}.

\bibitem[{Wang et~al.(2023)Wang, Wei, Schuurmans, Le, Chi, Narang, Chowdhery,
  and Zhou}]{cot_sc}
Xuezhi Wang, Jason Wei, Dale Schuurmans, Quoc~V Le, Ed~H. Chi, Sharan Narang,
  Aakanksha Chowdhery, and Denny Zhou. 2023.
\newblock Self-consistency improves chain of thought reasoning in language
  models.
\newblock In \emph{ICLR}.

\bibitem[{Wei et~al.(2022)Wei, Wang, Schuurmans, Bosma, Ichter, Xia, Chi, Le,
  and Zhou}]{cot}
Jason Wei, Xuezhi Wang, Dale Schuurmans, Maarten Bosma, Brian Ichter, Fei Xia,
  Ed~Chi, Quoc~V Le, and Denny Zhou. 2022.
\newblock Chain-of-thought prompting elicits reasoning in large language
  models.
\newblock In \emph{NeurIPS}, pages 24824--24837.

\bibitem[{West et~al.(2009)West, Pineau, and Precup}]{Wikispeedia}
Robert West, Joelle Pineau, and Doina Precup. 2009.
\newblock Wikispeedia: An online game for inferring semantic distances between
  concepts.
\newblock In \emph{IJCAI}, page 1598–1603.

\bibitem[{Yang et~al.(2024)Yang, Yu, Zhang, Cao, Xu, Zhang, Gonzalez, and
  Cui}]{bot}
Ling Yang, Zhaochen Yu, Tianjun Zhang, Shiyi Cao, Minkai Xu, Wentao Zhang,
  Joseph~E. Gonzalez, and Bin Cui. 2024.
\newblock \href {https://openreview.net/forum?id=ANO1i9JPtb} {Buffer of
  thoughts: Thought-augmented reasoning with large language models}.
\newblock In \emph{NeurIPS}.

\bibitem[{Yao et~al.(2022)Yao, Chen, Yang, and Narasimhan}]{webshop}
Shunyu Yao, Howard Chen, John Yang, and Karthik Narasimhan. 2022.
\newblock Webshop: Towards scalable real-world web interaction with grounded
  language agents.
\newblock In \emph{NeurIPS}, pages 20744--20757.

\bibitem[{Yao et~al.(2024)Yao, Yu, Zhao, Shafran, Griffiths, Cao, and
  Narasimhan}]{tot}
Shunyu Yao, Dian Yu, Jeffrey Zhao, Izhak Shafran, Tom Griffiths, Yuan Cao, and
  Karthik Narasimhan. 2024.
\newblock Tree of thoughts: Deliberate problem solving with large language
  models.
\newblock \emph{NeurIPS}, 36.

\bibitem[{Yao et~al.(2023)Yao, Zhao, Yu, Du, Shafran, Narasimhan, and
  Cao}]{react}
Shunyu Yao, Jeffrey Zhao, Dian Yu, Nan Du, Izhak Shafran, Karthik~R Narasimhan,
  and Yuan Cao. 2023.
\newblock {ReAct: Synergizing Reasoning and Acting in Language Models}.
\newblock In \emph{ICLR}.

\bibitem[{Zhilin et~al.(2018)Zhilin, Peng, Saizheng, Yoshua, William, Ruslan,
  and Manning}]{hotpotqa}
Yang Zhilin, Qi~Peng, Zhang Saizheng, Bengio Yoshua, Cohen William,
  Salakhutdinov Ruslan, and Christopher~D. Manning. 2018.
\newblock \href {https://doi.org/10.18653/v1/d18-1259} {Hotpotqa: A dataset for
  diverse, explainable multi-hop question answering}.
\newblock \emph{Proceedings of the 2018 Conference on Empirical Methods in
  Natural Language Processing}.

\bibitem[{Zhou et~al.(2024)Zhou, Yan, Shlapentokh-Rothman, Wang, and
  Wang}]{lats}
Andy Zhou, Kai Yan, Michal Shlapentokh-Rothman, Haohan Wang, and Yu-Xiong Wang.
  2024.
\newblock Language agent tree search unifies reasoning, acting, and planning in
  language models.
\newblock In \emph{ICML}.

\bibitem[{Zhou et~al.(2023)Zhou, Sch{\"a}rli, Hou, Wei, Scales, Wang,
  Schuurmans, Cui, Bousquet, Le, and Chi}]{least_to_most}
Denny Zhou, Nathanael Sch{\"a}rli, Le~Hou, Jason Wei, Nathan Scales, Xuezhi
  Wang, Dale Schuurmans, Claire Cui, Olivier Bousquet, Quoc~V Le, and Ed~H.
  Chi. 2023.
\newblock \href {https://openreview.net/forum?id=WZH7099tgfM} {Least-to-most
  prompting enables complex reasoning in large language models}.
\newblock In \emph{ICLR}.

\end{thebibliography}

\appendix
\section{Methods}
\label{sec:appendix-methods}

\xhdr{Input-Output (IO)} The simplest prompting style which uses the LLM to directly generate an output, with no intermediate steps.

\xhdr{Chain-of-Thought (CoT)} Solves the problem step by step by decomposing it into a sequence of thoughts \cite{cot}.

\xhdr{Reflexion} Generates linguistic feedback that is utilized during subsequent runs \cite{reflexion}.

\xhdr{Tree-of-Thoughts (ToT)} Decomposes the problem into multiple chain of thoughts, organized in a tree structure. Thought evaluation and search traversal algorithms are utilized to solve the problem \cite{tot}.

\section{Additional results}
\label{app:additional_results}
\begin{figure*}[!htb]
  \centering
  \begin{minipage}{\textwidth}
    \centering
    \includegraphics[width=\textwidth]{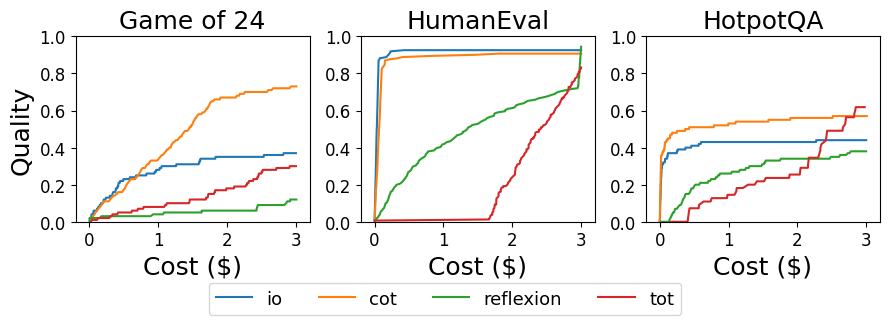}
    \figcaption{Comparing the cost-quality trade-off of IO, CoT, ToT, and Reflexion using \textbf{Llama-3.3-70B} as the base model. Within the indicated budget, simpler methods have similar or better performance complex ones while remaining cost-efficient.}
    \label{fig:cost_llama}
  \end{minipage}
\end{figure*}

\xhdr{Cost analysis}
In Fig.~\ref{fig:cost_llama}, we assess the cost-effectiveness of various prompting methods using the LLaMA 3.3 70B model. Similar to the results observed with GPT-4o-mini, both IO and CoT prompting strategies demonstrate significantly higher cost-efficiency. Notably, CoT maintains a substantial lead in tasks such as Game of 24 and HumanEval. However, for HotpotQA, the ToT approach slightly outperforms CoT.

\begin{figure*}[!htb]
    \centering
    \includegraphics[width=0.99\textwidth]{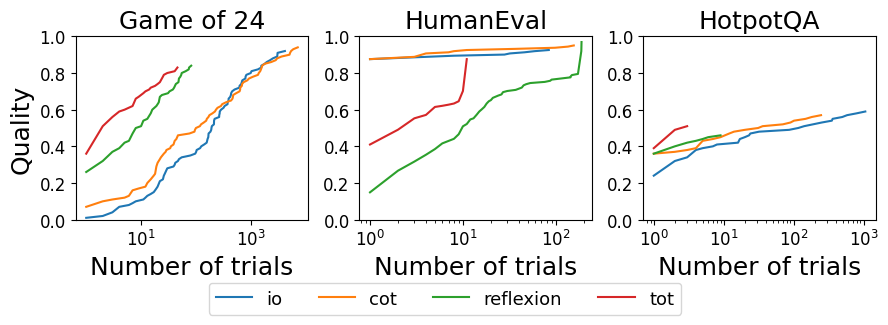}
    \figcaption{Comparing the sample-quality trade-off of IO, CoT, ToT, and Reflexion using \textbf{GPT-4o-mini} as the base model. Within the indicated budget, simpler methods outperform more complex ones while they remain sample-efficient only for the case of the HumanEval task.}    
    \label{fig:retrial_gpt}
\end{figure*}

\begin{figure*}[!htb]
    \centering
    \includegraphics[width=0.99\textwidth]{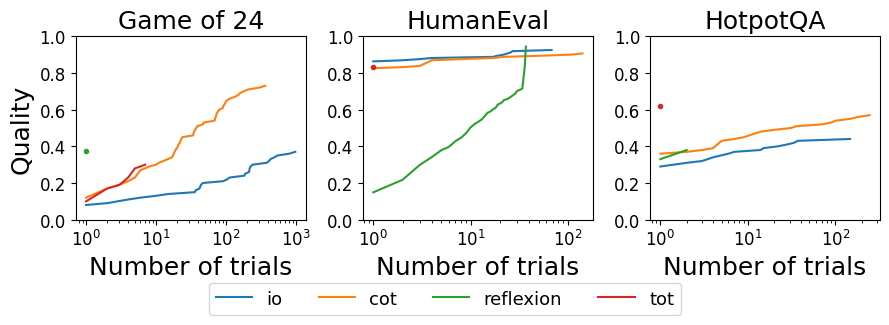}
    \figcaption{Comparing the sample-quality trade-off of IO, CoT, ToT, and Reflexion using \textbf{Llama-3.3-70B} as the base model. Within the indicated budget, simpler methods have better or similar performance than complex ones while they remain sample-efficient only for the case of the HumanEval task.}    
    \label{fig:retrial_llama}
\end{figure*}

\xhdr{Retrial analysis}. 
In Figs.~\ref{fig:retrial_gpt} and~\ref{fig:retrial_llama}, we evaluate the performance of each method as a function of the number of re-trials, under a fixed budget constraint. For Game of 24 and HotpotQA, ToT and Reflexion exhibit greater sample efficiency, achieving strong performance with relatively few re-trials. However, IO and CoT ultimately outperform them as the number of re-trials increases. In contrast, for HumanEval, IO and CoT are already more sample-efficient from the outset.

\begin{figure}[!htb]
    \centering
    \includegraphics[width=0.5\textwidth]{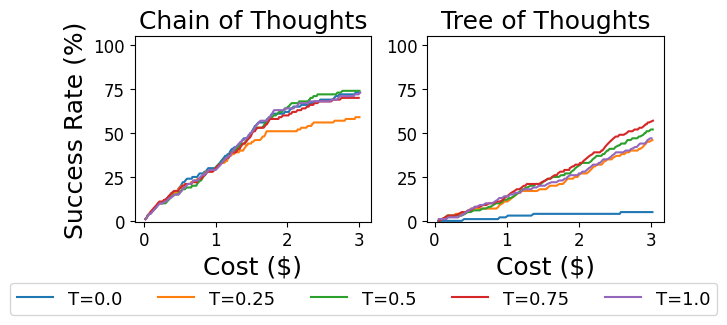}
    \figcaption{Comparing the cost-quality trade-off of CoT and ToT, using \textbf{Llama-3.3-70B} as the base model, across different temperature levels.}
    \label{fig:temp_llama}
\end{figure}

\xhdr{Temperature analysis}
In Fig~.\ref{fig:temp_llama}, we present the cost-effectiveness of CoT and ToT prompting for LLaMA 3.3 70B across varying temperature settings, under the same constrained budget as before. Unlike the results observed with GPT-4o-mini (Fig.\ref{fig:temp_gpt}), the performance trends here have not yet plateaued. We hypothesize that this is due to LLaMA 3.3 70B being approximately three times more expensive than GPT-4o-mini, placing the experiment in its early stages and limiting the conclusiveness of the results. In future work, we plan to increase the budget across all experiments to further investigate this behavior.

\section{Implementation Details}
\label{app:implementation_details}

\xhdr{Platforms}
GPT models were were accessed through the \href{https://platform.openai.com/docs/overview}{OpenAI API} while thhe utilization of the Llama models was facilitated by the  \href{https://docs.together.ai/docs/introduction}{ TogetherAI API}.

\xhdr{Model checkpoints and prices}
To compute the costs of our experiments we used the current model prices indicated OpenAI and Together AI, accordingly to the model. The specific models snapshot we used, along with their respective prices are presented in Table \ref{tab:snapshot_prices}.

\begin{table*}[!htb]
\centering
\begin{tabular}{|l|c|c|}

\hline
                            & US\$ per 1M prompt tokens & US\$ Per 1M completion tokens \\ \hline
\textbf{gpt-4o-mini} & 0.15               & 0.60                   \\ \hline
\textbf{LLaMA-3.3-70B}         & 0.88                 & 0.88                    \\ \hline
\end{tabular}
\centering
\mycaption{Model snapshot prices}{OpenAI and TogetherAI prices for each model used, during the implementation of the project.}
\label{tab:snapshot_prices}
\end{table*}

\end{document}